\documentclass[a4paper]{article}
\def\x{{\mathbf x}}
\def\y{{\mathbf y}}
\def\Y{{\mathbf Y}}
\def\p{{\mathbf p}}
\def\W{{\mathbf W}}
\def\L{{\mathcal L}}
\usepackage{amsmath}
\usepackage{INTERSPEECH2019}

\title{Self-supervised speaker embeddings
}
\name{Themos Stafylakis$^{1*}$, Johan Rohdin$^{2*}$, Old\v{r}ich Plchot$^2$, Petr Mizera$^1$, Luk\'a\v{s} Burget$^2$}
\address{
$^1$Omilia - Conversational Intelligence, Athens, Greece\\
\texttt{\{tstafylakis,petr.mizera\}@omilia.com} \\
$^2$Brno University of Technology, Faculty of Information Technology, IT4I Centre of Excellence, Czechia \\
\texttt{\{rohdin,iplchot,burget\}@fit.vutbr.cz} \\
\thanks{*Equal Contribution. \newline
{\bf Preprint. Work in progress.} \newline
 This project has received funding from the European union's Horizon 2020 research and innovation programme under the Marie Sklodowska-Curie and it is co-financed by the South Moravian Region under grant agreement No. 665860. The project was also supported by the Czech Science Foundation under project No. GJ17-23870Y.
}
}

\begin{document}

\maketitle
\begin{abstract}
  Contrary to i-vectors, speaker embeddings such as x-vectors are incapable of leveraging unlabelled utterances, due to the classification loss over training speakers. In this paper, we explore an alternative training strategy to enable the use of unlabelled utterances in training. We propose to train speaker embedding extractors via reconstructing the frames of a target speech segment, given the inferred embedding of another speech segment of the same utterance. We do this by attaching to the standard speaker embedding extractor a decoder network, which we feed not merely with the speaker embedding, but also with the estimated phone sequence of the target frame sequence.
  
  The reconstruction loss can be used either as a single objective, or be combined with the standard speaker classification loss. In the latter case, it acts as a regularizer, encouraging generalizability to speakers unseen during training. In all cases, the proposed architectures are trained from scratch and in an end-to-end fashion. We demonstrate the benefits from the proposed approach on VoxCeleb and Speakers in the wild, and we report notable improvements over the baseline.  
\end{abstract}

\section{Introduction}
During the last years, deep learning classifiers and representations have surpassed the performance of shallow and fully probabilistic counterparts in several tasks of speech recognition and computer vision, often by a large margin. A key ingredient towards this success was the availability of large annotated datasets, which enabled very deep architectures to be trained using supervised learning approaches. The availability of large in-domain corpora played a major role in building robust speaker recognition models, too. The success of Joint Factor Analysis and i-vectors can largely be attributed to such corpora, which enabled modeling correlations between acoustic units \cite{kenny2007joint,dehak2011front}. More recently, deep learning architectures outperformed such methods in most of the speaker recognition benchmarks \cite{snyder2018x}.

On the other hand, these architectures require the datasets to be labelled with respect to speaker, which was not the case with i-vectors. As an unsupervised model, an i-vector extractor does not require utterances with associated speaker labels for training \cite{kenny2007joint}. Labelled utterances are needed merely for training the backend classifier (typically a PLDA) which requires much less data, thanks to its relatively small number of trainable parameters.

In this paper, we introduce a training architecture capable of learning speaker embeddings with only few or no speaker labels. The structure we add to the standard speaker embedding network is a decoder network, which learns how to reconstruct speech segments in the frame-level, using a mean-squared error loss. A key idea of the method is the conditioning of the decoder not merely on the embedding extracted by the encoder (i.e. the embedding extractor), but also on the phonetic sequence of the decoding speech segment, as estimated by an independently trained Automatic Speech Recognition (ASR) model. Such a conditioning allows for a decoding loss over speech frames to apply for learning in an end-to-end fashion using standard backpropagation. It moreover enables learning speaker embeddings that capture only the idiosyncratic characteristics of a speaker, rather than irrelevant information about the phonetic sequence. The latter property is further improved by extracting two different segments from an utterance: the first for feeding the encoder and extracting the embedding, and the second one for using it as target for the decoder, together with its associated phone sequence. 

We show that the proposed decoder loss can be combined with the standard x-vector architecture and loss (i.e. cross-entropy over training speakers) yielding significant improvement. Finally, we consider a semi-supervised learning scenario, where only a small fraction of the training utterances contain speaker labels and we show how the proposed architecture can leverage both labelled and unlabelled utterances. All our experiments are conducted on VoxCeleb and Speakers In The Wild benchmarks.

\section{Related work}
\subsection{Speaker recognition using autoencoders} There have been several attempts in speaker recognition to make use of reconstruction losses. Most of them are based on (plain or variational) autoencoders, either in an unsupervised way or using speaker labels \cite{chien2017variational,villalba2017tied,SilnovaIS18}. Other such approaches aim at reducing the phonetic variability of short segments by learning a mapping from short segments to the whole utterance. The main weakness of these methods is the fact that they operate over fixed, utterance-level representations, typically i-vectors \cite{guo2018deep}. Our approach of conditioning the reconstruction on the estimated phone sequence of each segment can be employed, enabling such approaches to be revisited in an end-to-end fashion. Other recent approaches aiming at enhancing the x-vector architecture with adversarial loss are also relevant, since they are propose joint training of the network with auxiliary losses and structures which are removed in runtime \cite{ding2018mtgan,rohdin2018speaker,bhattacharya2018generative}. 

\subsection{Speech synthesis, recognition, and factorization} Recently, speaker embeddings have been deployed in text-to-speech (TTS) and voice conversion \cite{gibiansky2017deep,chen2018sample,jia2018transfer}. The embeddings are typically extracted using a pretrained network (e.g. a d- or x-vector extractor), which may be fine-tuned to the task \cite{chen2018sample}. Conditioning the decoder on speaker embeddings (together with the text of the target utterance) is crucial for training multispeaker TTS systems and producing synthetic speech for target speakers unseen during training. Although our method shares certain similarities with this family of TTS methods (especially in the decoder), our goals are different. Rather than employing a pretrained speaker recognition model to extract embeddings, we demonstrate that speaker-discriminative training is feasible using merely a reconstruction loss over speech segments and training the overall network jointly. Finally, a recently introduced ASR approach for integrating ASR and TTS into a single cycle during training has also certain similarities with our method (\cite{tjandra2018end}), and the same holds for deep factorization method for speech proposed in \cite{li2018deep}.

\subsection{Self-supervised learning}
The approach of extracting speaker embeddings via reconstructing different parts of a sequence can be consider as an application of self-supervised learning, where a network is trained with a loss on a pretext task, without the need for human annotation. Models using self-supervised learning for initialization are now state-of-the-art in several domains and tasks, such as action recognition, reinforcement learning, and natural language understanding \cite{wiles2018self,sermanet2018time,noroozi2018boosting,devlin2018bert}. 

\section{The proposed architecture}
\begin{figure}[!htbp]
\centering
\includegraphics[clip, trim=4.0cm 0cm 0cm 0cm, width=4.0in]{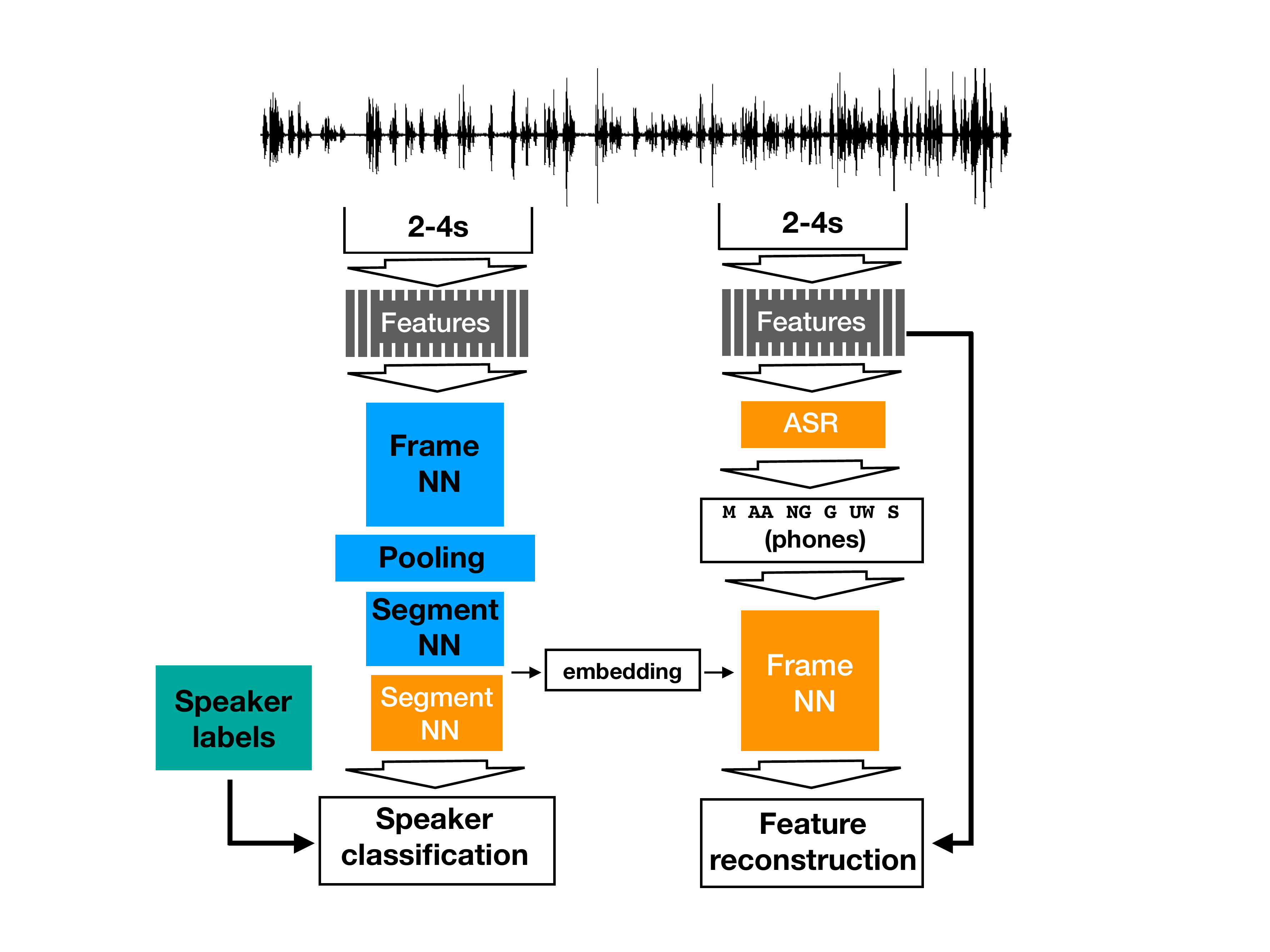}
\caption{\label{fig:arch} Block-diagram of the architecture. Note that only the blue part of the network is required in runtime.}
\label{diagram}
\end{figure}

In this section we describe the network used in training and we provide rationale for certain algorithmic choices. The architecture is depicted in Figure \ref{fig:arch}. Architectural details are given in Section \ref{ss:imp_det}. 
\subsection{Notation}
We denote by $\Y = \{\y_t\}_{t=1}^T$ the frame sequence of an utterance, and the corresponding estimated phones sequence by $\hat{\p} = \{\hat{p}_t\}_{t=1}^T$. We also denote an associated version of the same utterance by $\tilde{\Y} = \{\tilde{\y}_t\}_{t=1}^T$, where tilde indicates corruption by noise, reverberation, or another data augmentation scheme. The encoder consumes randomly extracted segments from $\tilde{\Y}$, denoted by $\tilde{\Y}_{e} = \{\tilde{\y}_t\}_{t=t_e}^{t_e+\tau_e-1}$, where $\tau_e$ is randomly sampled in $[200,400]$. Let also $\Y_{d} = \{{\y}_t\}_{t=t_d}^{t_d+\tau_d-1}$ be another segment of the same utterance (without data augmentation), and let $\hat{\p}_{d}$ denote its corresponding estimated phone sequence. The encoder part of the network (i.e. the embedding extractor) is denoted by
\begin{equation}
\x_{e} = f_{e}(\tilde{\Y}_{e};\W_e),    
\end{equation}
and it is a function parametrized by $\W_e$. 
\subsection{Purely self-supervised training}
We train the network using a decoder, which is a network implementing the following function
\begin{equation}
\hat{\Y}_{d} = f_{d}(\x_{e},\hat{\p}_{d};\W_d).
\end{equation}
The decoder is parametrized by $\W_d$ and receives as input the embedding $\x_{e}$ and the estimates phone sequence $\hat{\p}_{d}$. Note that the embedding $\x_{e}$ and target frames $\Y_{d}$ correspond in general to different segments of the same utterance.
The architecture is trained using mean-squared error (MSE) loss, i.e.
\begin{equation}
\L_{MSE}(\Y_{d},\hat{\Y}_{d}) = \frac{1}{\tau_d}\sum_{t=t_d}^{t_d+\tau_d-1} \left\lVert \y_t - \hat{\y}_t \right\rVert ^2.
\end{equation}
The above equations show that the encoder and decoder can be trained jointly using standard backpropagation.
\subsection{Combining self-supervision with cross-entropy over speakers}
In case the utterance $\Y$ has a speaker label $s$ we may combine the decoding loss with the standard cross-entropy loss over speakers. The speaker classifier estimates the posterior distribution over the set of training speakers, i.e. 
\begin{equation}
P(s | \x_{e},\W_c) = f_{c}(\x_{e};\W_c),
\end{equation}
it is parametrized by $\W_c$ and has a softmax as final layer. The cross-entropy (CE) loss $\L_{CE}(s,P(s | \x_{e},\W_c))$ can be added to the decoding loss and the overall network can be trained jointly, i.e.
\begin{equation}
\L_{joint}(\cdot) = \delta\L_{CE}(\cdot) + \alpha \L_{MSE}(\cdot),
\end{equation}
where $\delta = \{0,1\}$ indicates whether the particular utterance has a speaker label, and $\alpha \geq 0$ is a scalar for balancing the two losses.
\subsection{Discussion}
\subsubsection{Encoding and decoding segments}
The rationale for defining the encoding and decoding sequences as different segments of the same utterance is to encourage the encoder to learn embeddings that encode information about the way a speaker pronounces acoustic events unseen in the encoding sequence. Note also that we keep the decoding target sequence clean (i.e. without augmentation). We do so for encouraging the encoder to learn how to denoise speech sequences, an approach similar to denoising autoencoders \cite{vincent2008extracting}.

\subsubsection{Representation of the phonetic sequence}
For passing the phonetic sequence to the decoder, we choose to define estimated phones as phonetic units, in the form of one-hot vectors. Clearly, there are several other options, such as bottleneck features, senones or characters. 

One of the reasons why we did not use bottleneck features is that they inevitable carry speaker-discriminative information, as the experience shows (recall that speaker recognition is feasible with plain bottleneck features \cite{lozano2016analysis}). Therefore, passing bottleneck features to the decoder could create information leakage, preventing the embedding from capturing useful speaker-discriminative information. Another drawback of bottlenecks is that they would tie the remaining network to a specific bottleneck extractor. Contrarily, symbolic entities such as phones allow for different ASR models to be used for estimating $\hat{\p}_{d}$ or even for using the ground truth phones, when available. 

On the other hand, senones would result in a much larger and harder to train decoding network, while the senone posteriors would be much less spiky compared to phones, and hence far from resembling one-hot vectors. Furthermore, passing senones to the decoder seems unnecessary; the decoder can recover the context-dependence of each phone since it is conditioned on the overall phone sequence $\hat{\p}_{d}$. 

Finally, using characters would require additional complexity to align the two sequences, such as an attention mechanism employed in TTS approaches \cite{chen2018sample}. For these reasons, we consider phones in the form of one-hot vectors as the appropriate representation and level of granularity for this task and setup.

\subsubsection{Semi and weakly supervised learning}
The proposed architecture defines a principled way of leveraging unlabelled data in x-vector training. There are other losses for supervised training with which one may combine it, such as the triplet loss
\cite{schroff2015facenet,zhang2017end,li2017deep}. Note that both cross-entropy and triplet losses cannot leverage unlabelled utterances (e.g. by splitting the same utterances into multiple segments), unless one assumes that each utterance in the training set is coming from a different speaker (which is typically not the case). On the other hand, our self-supervised method requires only the knowledge that two segments belong to the same speakers, while it can be extended to encode and decode on segments coming different utterances of the same speaker. This would make it suitable also for certain weakly supervised learning settings, where labels indicate merely that two or more utterances are coming from the same speaker, without excluding the possibility that other utterances may belong to the same speaker as well. In such cases, a training criterion that does not require exclusive labels (cross-entropy loss) or negative pairwise labels (triplet loss) seems to be the only principled method for learning speaker representations.

\section{Experiments}

\subsection{VoxCeleb and SITW datasets}
We evaluate the systems on the Speakers in the wild (SITW) \cite{McLaren_2016} \emph{core-core eval} set and the VoxCeleb 1 \emph{test} set \cite{Nagrani17}. We use the SITW core-core development set for tuning various hyperparamters of the systems. For preparing the data, we use the Kaldi \cite{povey2011kaldi} SITW recipe ({\texttt{sitw/v2}}). This recipe uses VoxCeleb 1 and 2 \cite{Chung18b} for training data. We use the recipe as is, except that we do not include VoxCeleb 1 test set in the training set. The number of speakers in the training set is 7146 and the number of utterances is 2081192 including augmentations. For \emph{semisupervised} experiments, we randomly selected 1000 speakers, having in total 227998 utterances.

\subsection{Implementation and training details}
\label{ss:imp_det}
\subsubsection{Implementation and decoder}
We use the TensorFlow toolkit \cite{tensorflow2015-whitepaper} for implementing our systems. As baseline, we use the standard Kaldi x-vector architecture \cite{snyder_interspeech_2017}, i.e. five TDNN layers with ReLU activation functions followed by batch normalization, followed by a pooling layer that accumulates mean and standard deviations, followed by two feed-forward layers with ReLU and batch normalization, and finally a softmax layer for classifying speakers. Different from Kaldi, we apply a global normalization on the input features and batch normalization also after the pooling layer. As discussed above, the loss is CE over training speakers.

The reconstruction network (i.e. the decoder) consists of five layers that operates framewise. Its input are the phone labels represented as one-hot vector and its output predicts the 30-dimensional feature vectors. The input layer is either (a) a feed-forward layer or (b) a TDNN layer with a context of three frames on each side (denoted by ctx). The other layers are feed-forward layers with an output dimension 166 (i.e. same as the number of phone labels). All layers except the last one are followed by ReLU and batch normalization. The embedding is appended to the input of each layer. The loss for the reconstruction is the MSE between the real and predicted features. 

In the experiments, we use minibatches containing 150 segments. The lengths of the segments are 2-4s. We use the ADAM optimizer \cite{DBLP:journals/corr/KingmaB14}, starting with a learning rate of 1e-2 which we then halve whenever the loss on a validation set does not improve for 32 epochs, where an epoch is defined to be 400 minibatches. In the semisupervised experiment, each batch contains 150 labelled segments and 150 unlabelled segments and only the labelled segments will be used to calculate the speaker classification loss.

\subsubsection{The ASR model}
The frame-level phone labels are generated using the official Kaldi \cite{povey2011kaldi} Tedlium speech recognition recipe ({\texttt{s5\_r3}}).  This recipe uses a TDNN based acoustic model with i-vector adaptation and an RNN based language model. Phone posteriors are obtained from the lattices using the forward-backward algorithm and then converted to hard labels. There are 39 phones, each coming in four different versions depending on their position in the word, plus a silence (SIL) and noise class (NSN) that has 5 versions each, resulting in 166 \emph{phone classes}.
\subsubsection{PLDA Backend}
%
We used an identical backend to the one in the Kaldi x-vector recipe. This backend involves a preprocessing step which first reduces the x-vector dimension by LDA from 512 to 128, and then applies a nonstandard variant of length-norm\footnote{https://github.com/kaldi-asr/kaldi/blob/master/src/ivector/plda.cc}. The backend was implemented in python based on our in-house toolkit {\it Pytel}. For the fully supervised experiments we, as the Kaldi recipe, use the 200k longest utterances, resulting in 6298 speakers. For the semi-supervised experiments, we use all of these utterances where the speaker is among the 10000 randomly selected, resulting in 899 speakers and 31785 utterances.

\subsection{Experimental Results}
\subsubsection{Fully labelled training set}
The results using the standard training set of VoxCeleb are given in Table \ref{tab:Full}.  
\begin{table}[!ht]
\caption{\label{tab:Full} Results with fully supervised training. EER refers to \emph{Equal Error Rate} in \% and mDCF refers to \emph{Minimum Detection Cost} with $\mathrm{P}(\mathrm{tar})=0.01$. All $N_f \approx $7k speakers are used in PLDA training. Baseline refers to the standard x-vector recipe.}
    \begin{tabular}{l r r r r } 
    \hline
     & \multicolumn{2}{c}{SITW} &   \multicolumn{2}{c}{VoxCeleb} \\ 
     & EER  & mDCF & EER  & mDCF \\
    \hline
    \hline
     Spk, baseline   & 3.879	& 0.382  & 4.088 & 0.431  \\ 
     \cline{1-5}
     Self             & 5.823	& 0.540 & 6.707	& 0.628 \\  
     Self, cln      & 5.685	& 0.542 & 6.538	& 0.612 \\
     Self, cln, ctx &  4.844	& 0.485 & {\bf 5.758}	& {\bf 0.536} \\
     Self, cln, ctx, same  &  {\bf 4.760} & {\bf 0.478} & 6.347 &	0.580  \\
     \hline
     Spk+Self         & {\bf 3.311}	& 0.362 & 3.961	& 0.386  \\
     Spk+Self, cln  &   3.362 	& 0.356 & 3.881	& 0.380 \\ 
     Spk+Self, cln, ctx &  3.362 	 &  {\bf 0.353}  & {\bf 3.759 }& {\bf 0.344} \\
     \hline
    \hline 
   \end{tabular}
  \vspace{-0.2cm}
\end{table}

The results show that the self-supervised models are capable of extracting speaker-discriminant embeddings. The use of clean (cln) decoding utterances yields slightly better results, as it enforces the encoder to act as a denoiser. Moreover, conditioning the decoding TDNN on a 7-frame phonetic context (ctx) is clearly beneficial compared to conditioning it merely on the phone of the target frame. 

We also observe that using same segments for encoding and decoding (same) yields inferior performance on VoxCeleb, while on SITW their performance is equivalent. A plausible explanation is that VoxCeleb contains shorter segments compared to SITW. Hence, as encoding and decoding on different segments encourages the network to learn how to reconstruct phonetic subsequences unseen in the encoding segments, it is expected to be more beneficial for short durations.    

When the two losses are combined (Spk+Self), the model clearly outperforms plain x-vector (Spk). In this case, the self-supervised loss has a regularization effect, constraining the network to learn representations that generalize well to unseen speakers. Again, the use of context yields superior performance, although in this case the differences are less significant.

\subsubsection{Partly labelled training set}
In this set of experiments, we assume that only a fraction of the training utterances is labelled. Hence, in the results we provide in Table \ref{tab:Semi} the PLDA is trained with $N_r$ = 899 VoxCeleb speakers out of $N_f \approx 7K$.  
\begin{table}[!ht]
\caption{\label{tab:Semi} Results with semi-supervised training. EER refers to \emph{Equal Error Rate} in \% and mDCF refers to \emph{Minimum Detection Cost} with $\mathrm{P}(\mathrm{tar})=0.01$. In all experiments, the same set of $N_r$ = 899 randomly selected speakers is used for PLDA training. Baseline refers to the standard x-vector recipe.}
    \begin{tabular}{l r r r r } 
    \hline
     & \multicolumn{2}{c}{SITW} &   \multicolumn{2}{c}{VoxCeleb} \\ 
     & EER  & mDCF & EER  & mDCF \\
    \hline
    \hline
     Spk, baseline    &  6.616 &	0.593 & 7.709 & 0.658 \\ 
     Spk, full set &  4.347 & 0.424 & 4.804 & 0.472  \\ 
     \hline     
     Self, cln            & 6.748 &	0.615 & 7.619 &	0.668    \\
     Self, cln, ctx     & 5.793 &	{\bf 0.548} & {\bf 6.644} &	{\bf 0.589}    \\
     Self, cln, ctx, same   &  {\bf 5.768}	& {\bf 0.548} & 7.503	& 0.612  \\
     \hline
     Spk+Self, cln        & 5.715	& 0.497 & 6.972	& 0.543   \\ 
     Spk+Self, cln, ctx & {\bf 5.416} & {\bf 0.488} & {\bf 6.315} & {\bf 0.534} \\
     \hline
    \hline 
   \end{tabular}
  \vspace{-0.2cm}
\end{table}

In the first two experiments in Table \ref{tab:Semi} we use standard CE over speakers loss. We observe the severe degradation when the number of speakers used to train the x-vector baseline is reduced to $N_r = 899$ (Spk, baseline). For comparison, we report the experimental results where the full set of speakers is used for training the x-vector model (Spk, full set). 

The results using only self-supervision with context are clearly superior to those of pure x-vectors, due to the capacity of self-supervision in leveraging all available utterances during training. Moreover, when the two losses are combined, the results become even better, especially in terms of minDCF. Finally, we observe again the gains in performance by using different encoding and decoding segments.

\section{Conclusions and future work}

In this paper, we introduced a new way of training speaker embeddings extractors using self-supervision. We showed that a typical TDNN-based extractor can be trained without speaker labels, using a decoder network to approximate in the MSE sense a speech segment of the same utterance. A key idea for enabling decoding is the conditioning of the decoder on both the embedding and the phonetic sequence of the decoding segment, as estimated by an ASR model. Furthermore, we showed that the proposed loss can be combined with the standard cross-entropy, yielding notable improvements. Finally, we demonstrated its effectiveness on semi-supervised learning, i.e. when only a small fraction of the training set is labelled. Both additional networks we introduced (decoder and ASR model) are only needed during training, leaving the standard x-vector architecture unchanged in runtime.

The proposed approach can be extended in several ways. The method of conditioning the decoder on the phonetic sequence of the speech segment paves the way for revisiting methods such as variational autoencoders in an end-to-end fashion. Speech synthesis approaches may also benefit from the proposed method, e.g. by training embedding extractors jointly with TTS from scratch. Finally, there is large room for improvement in the architecture (e.g. by using a recurrent or attentive decoder or deeper and wider encoder \cite{snyder2019speaker}), in the training scheme (e.g. by varying the duration of encoding and decoding segments), and in the way the existing speaker labels are used in training (e.g. by extracting the two segments from different utterances of the same speaker).
\bibliographystyle{IEEEtran}

\bibliography{mybib}


\end{document}